\documentclass[letterpaper, 10 pt, journal, twoside]{IEEEtran}

\IEEEoverridecommandlockouts           

\PassOptionsToPackage{hyphens}{url}\usepackage{hyperref}

\hypersetup{
    colorlinks=true,
    linkcolor=blue,
    filecolor=blue,      
    citecolor=blue,
}

\usepackage{math}
\usepackage{tablefootnote}

\newcommand{\spheading}[2][6.4em]{
  \rotatebox{90}{\parbox{#1}{\centering \small {\textbf{#2}}}}\hspace{0.1em}}

\newcommand{\thickline}{\tikz[baseline=-0.5ex]\draw[thick,line width=2pt](0,0)--(3.5ex,0);}

\usepackage{tcolorbox}
\definecolor{mycolor}{rgb}{0.811, 0.882, 0.952}
\newtcolorbox{mybox}{colback=gray!10!white,colframe=black!50!black, boxrule=0.4pt,arc=6pt, left=6pt,right=6pt,top=5pt,bottom=5pt}

\newcommand{\ie}{{\em i.e.,~}}

\newcommand{\ours}{VL-TGS}

\begin{document}

\title{\ours: Trajectory Generation and Selection using Vision Language Models in Mapless Outdoor Environments}

\author{Daeun Song$^{1}$$^{\star}$, Jing Liang$^{2}$$^{\star}$, Xuesu Xiao$^{1}$, and Dinesh Manocha$^{2}$

\thanks{$^{\star}$The majority of the work was conducted while Daeun Song was a postdoctoral researcher at the University of Maryland.}
\thanks{$^{\star}$Equal contribution.}

\thanks{$^{1}$D. Song and X. Xiao are with the Department of Computer Science, George Mason
University, Fairfax, VA 22030 USA (e-mail: dsong26@gmu.edu; xiao@gmu.edu).}
\thanks{$^{2}$J. Liang and D. Manocha are with the Department of Computer Science, University of Maryland, College Park, MD 20742 USA (e-mail: jingl@umd.edu; dmanocha@umd.edu).}


}


\maketitle

\begin{abstract}
We present a multi-modal trajectory generation and selection algorithm for real-world mapless outdoor navigation in human-centered environments. Such environments contain rich features like crosswalks, grass, and curbs, which are easily interpretable by humans, but not by mobile robots. 
We aim to compute suitable trajectories that (1) satisfy the environment-specific traversability constraints and (2) generate human-like paths while navigating on crosswalks, sidewalks, etc. Our formulation uses a Conditional Variational Autoencoder (CVAE) generative model enhanced with traversability constraints to generate multiple candidate trajectories for global navigation. We develop a visual prompting approach and leverage the Visual Language Model's (VLM) zero-shot ability of semantic understanding and logical reasoning to choose the best trajectory given the contextual information about the task. We evaluate our method in various outdoor scenes with wheeled robots and compare the performance with other global navigation algorithms. In practice, we observe an average improvement of $20.81\%$ in satisfying traversability constraints and $28.51\%$ in terms of human-like navigation in four different outdoor navigation scenarios. 
\end{abstract}


\section{Introduction}
\label{sec:intro}

\begin{figure}[t]
\centering
\includegraphics[width=\linewidth]{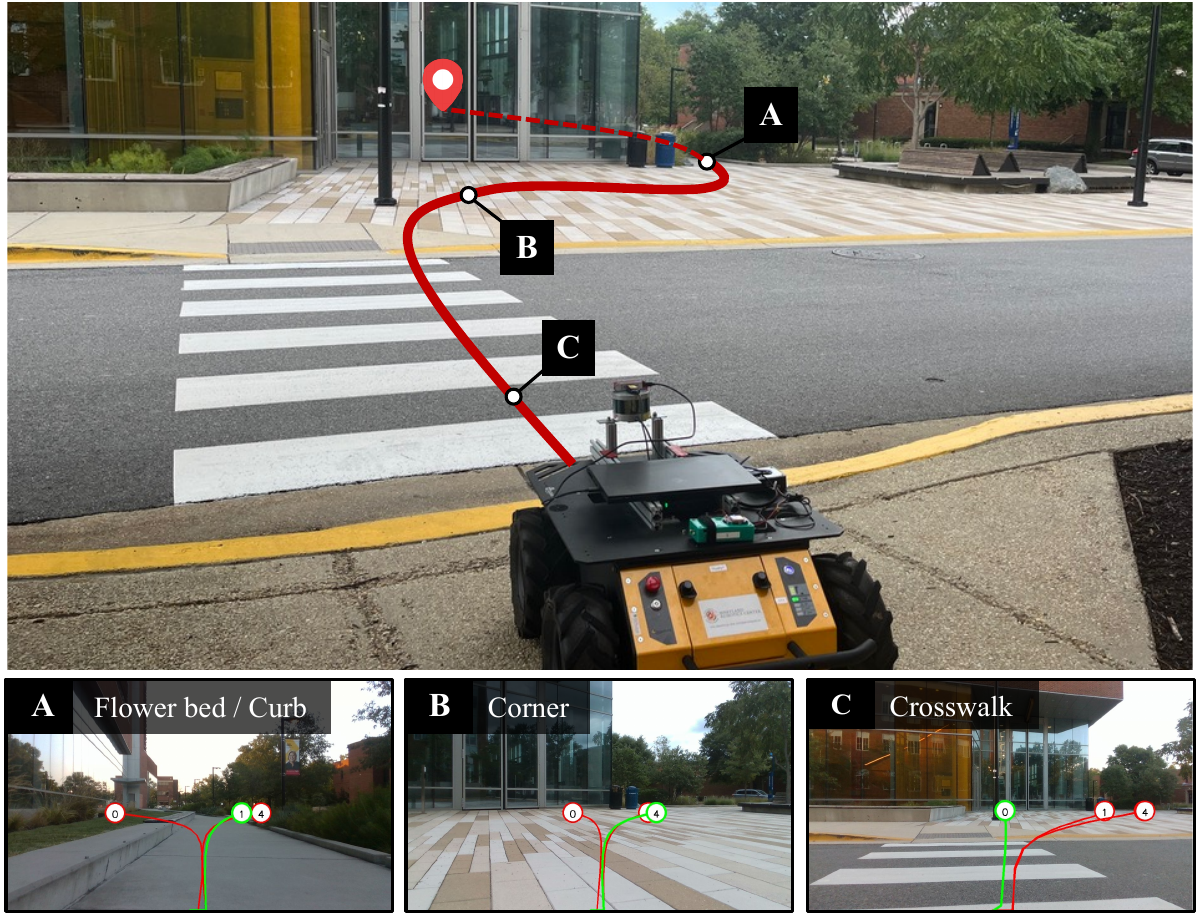} \vspace{-1.5em}
\caption{Trajectories generated and selected using \ours~in outdoor navigation. The example path includes three different types of scenarios: (A) flower bed and curb, (B) corner, and (C) crosswalk. On the top, the map pin icon marks the goal behind the building, with the red solid or dashed line highlighting the robot's path. On the bottom, candidate trajectories are marked in red lines with numbers. The green path corresponds to the trajectory computed using \ours. Overall, \ours~is capable of generating diverse, geometrically traversable paths and selecting semantically feasible trajectories for navigation in human-centered environments.} \vspace{-1.5em}
\label{fig:cover}
\end{figure}

\IEEEPARstart{M}{apless} outdoor navigation requires robots to compute trajectories or directions in large-scale environments without relying on pre-built maps. This problem is particularly important for global navigation in outdoor settings, where creating and maintaining accurate maps is impractical due to dynamic changes such as constructions~\cite{wijayathunga2023challenges, jeong2021motion}. Unlike map-based methods that depend on detailed geometric representations of the environment~\cite{ganesan2022global, gao2019global, psotka2023global}, mapless techniques rely directly on sensory input~\cite{mtg, liang2024dtg}, requiring robots to adapt to environmental changes and navigate through unknown spaces without the need for prior knowledge. 

Traditionally, both map-based and mapless navigation approaches have relied on traversability analysis based on geometric shapes, often using LiDAR data to identify navigable regions~\cite{mtg, liang2024dtg, suger2015traversability}. While this approach is effective for detecting larger obstacles and general terrain features, it faces challenges in nuanced environments~\cite{weerakoon2023graspe, sathyamoorthy2024mim}. Features such as short grass, curbs, and low-profile flower beds can be challenging for LiDAR to detect due to their subtle and low-profile characteristics. Additionally, while geometric environmental data is sufficient for navigation in obstacle-rich environments, it falls short in human-centered environments. 

Navigating human-centered outdoor environments requires advanced scene understanding to ensure safety and reliability~\cite{moller2021survey}. Robots must not only recognize physical features, such as walkways, crosswalks, and paved paths, but also interpret their intended use within the environment and navigate accordingly. For example, paved roadways may only be temporarily used when construction blocks the sidewalk, but they can always be used to cross a street when marked with a zebra crossing. This involves identifying areas designated for pedestrian movement, detecting obstacles or temporary changes, and understanding how these elements influence viable paths. Achieving this requires contextual reasoning to understand and adapt to the implicit rules and expectations of human-centered environments~\cite{charalampous2017recent}.

To build such contextual understanding of the environment, many existing methods~\cite{weerakoon2023graspe, zhang2022trans4trans} rely on segmentation or classification~\cite{kirillov2023segment, borges2022survey}. However, these require extensive training with ground truth data and are limited to labeled datasets. This limitation hinders their generalizability to unknown scenes. 
Recent advances in Large Language Models (LLMs) and Vision Language Models (VLMs) have demonstrated strong zero-shot capabilities across a wide range of tasks, including logical reasoning~\cite{austin2021program, ha2022semantic} and visual understanding~\cite{alayrac2022flamingo, radford2021learning}. VLMs, in particular, have the ability to process and understand both visual and textual information, enabling them to perform a wide range of multi-modal tasks. Their ability to reason contextually and adapt their outputs to align with implicit environmental rules makes them ideal for navigating human-centered outdoor spaces.

{\bf Main Results:} 
We present \ours, a novel multi-modal approach for trajectory generation and selection in mapless outdoor navigation (Fig.~\ref{fig:cover}). Our method combines LiDAR-based geometric information with RGB image data for comprehensive traversability analysis and scene understanding. Using a CVAE-based~\cite{cvae} approach, we first generate multiple candidate trajectories based on the LiDAR scene perception. A VLM is then employed for trajectory selection based on the environmental context understanding through RGB image data. While VLMs lack the capability to produce precise spatial outputs, they can effectively utilize visual annotations to guide the selection process among a discrete set of coarse options~\cite{shtedritski2023does, nasiriany2024pivot, yang2023set}. By incorporating VLMs, our approach enables human-like decision-making to select optimal trajectories from the candidates, ensuring they align with geometric traversability constraints while addressing the contextual demands of global navigation. 
We demonstrate the effectiveness of our approach in outdoor scenarios featuring diverse human-centered environments and navigation challenges, such as crossing streets at crosswalks and adhering to walkways. The major contributions of our work include: 
\begin{enumerate}
    \item A novel integrated trajectory generation and selection method, \ours, to generate multiple candidate trajectories using a CVAE-based approach and to select the most suitable trajectory using the VLM with visual prompting. Our CVAE-based trajectory generation method generates multiple candidate trajectories that are traversable considering the geometrical information retrieved from the LiDAR sensor. Our VLM-based trajectory selection method selects the best trajectory, which is traversable, in terms of both a geometric and semantic manner suitable for a human-centered environment.

    \item We explore the use of a visual prompting approach to enhance the spatial reasoning capabilities of VLMs in the context of trajectory selection. By incorporating visual markers such as lines and numerical indicators within the RGB image, we provide explicit guidance to the VLM. 
    We conduct ablation studies, first demonstrating the importance of providing high-quality candidate trajectories, and then comparing the effectiveness of having a visual marking method.

    \item We evaluate \ours~in four different outdoor scenarios. We measure the satisfaction rate of traversability constraints and the Fr{\'e}chet distance with respect to a human-{teleoperated} trajectory. We compare the results with state-of-the-art trajectory generation approaches. 
    We observe an average improvement of $20.81\%$ in the traversability satisfaction rate and $28.51\%$ in the Fr{\'e}chet distance. We also qualitatively demonstrate the benefits of our approach over other methods.
\end{enumerate}

\section{Related Work}

\label{sec:background}
This section reviews related works on outdoor robot navigation, with a particular focus on trajectory generation.

\subsection{Outdoor Robot Navigation}
Reinforcement-learning-based motion planning approaches~\cite{liang2021crowd, hao2023exploration} use an end-to-end structure to take observations and generate actions or trajectories. However, these methods are designed for short-range navigation, and on-policy reinforcement learning approaches also suffer from the reality gap. Map reconstruction with path planning approaches \cite{path_planning, zhai2022path} provides a solution for global planning by building a map during navigation, but these approaches require a large memory for the global map. To address this issue, NoMaD \cite{nomad} and ViNT \cite{vint} use topological maps to reduce memory usage for navigation, but these approaches require topological nodes to be predefined, making them unsuitable for fully unknown environments. To overcome these limitations, our approach uses a CVAE-based trajectory generation method~\cite{mtg} to generate trajectories and leverages VLMs to select the optimal trajectory to reach the goal.


\begin{figure*}[ht]
\centering
\includegraphics[width=0.7\linewidth]{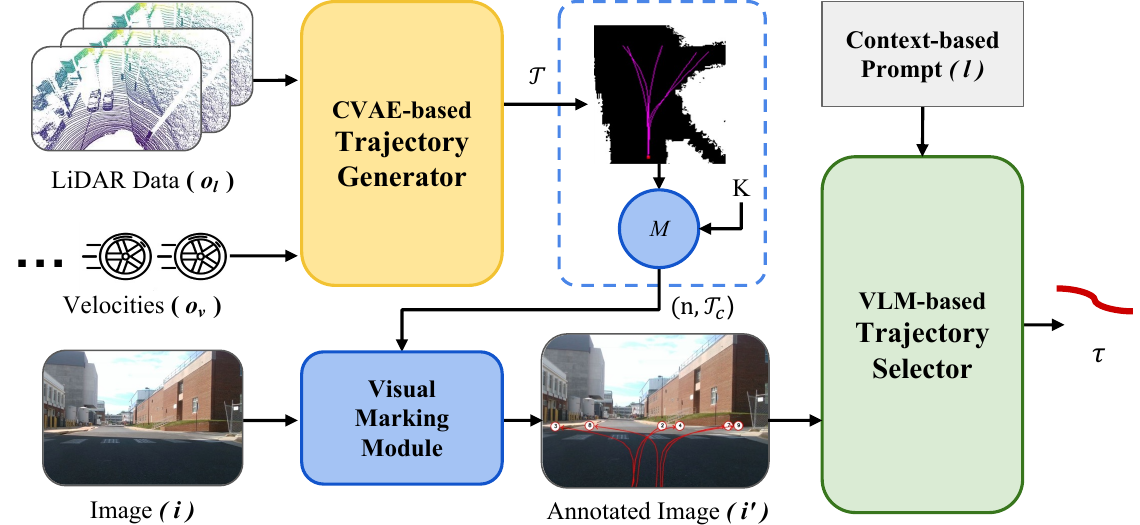}
\caption{\textbf{Architecture:} Our approach consists of two stages: CVAE-based trajectory generation and VLM-based trajectory selection. In the first stage, our attention-based CVAE takes consecutive frames of LiDAR point clouds and robot velocities as input, generating multiple diverse trajectories. These trajectories are sorted and visually marked with lines and numbers in the robot-view RGB image. In the second stage, our VLM-based trajectory selection module identifies the best trajectory number based on semantic feasibility, ensuring it lies on the sidewalk, avoids structures, crosses at zebra crossings, and adheres to other contextual rules.}\label{fig:architecture}
\vspace{-1em}
\end{figure*}

\subsection{Vision Language Models in Navigation}
Recent breakthroughs in Language Foundation Models (LFMs)~\cite {bommasani2022opportunities}, encompassing VLMs and LLMs, demonstrate significant potential for robotic navigation. LM-Nav~\cite{shah2023lm} employs GPT-3 and CLIP~\cite{radford2021learning} to extract landmark descriptions from text-based navigation instruction and ground them in images, effectively guiding a robot to the goal in outdoor environments. VLMaps~\cite{huang2023visual} propose a spatial map representation by fusing vision-language features with a 3D map that enables natural language-guided navigation. CoW~\cite{gadre2022clip} performs zero-shot language-based object navigation by combining CLIP-based maps and traditional exploration methods. Most of these researches focus on utilizing VLMs for high-level navigation guidance by extracting text-image scene representation. For low-level navigation behaviors, VLM-Social-Nav~\cite{song2024vlm} explores the ability of VLM to extract socially compliant navigation behavior with the interaction with social entities like humans. CoNVOI~\cite{convoi} uses visual annotation to extract a sequence of waypoints from camera observation to navigate robots. PIVOT~\cite{nasiriany2024pivot} uses visual prompting and optimization with VLMs in various low-level robot control tasks including indoor navigation. It shows the potential of a visual prompting approach for VLMs in robotic and spatial reasoning domains.

Building on these approaches, our work uses a VLM to guide low-level navigation behavior by understanding contextual and semantic information about the surroundings. We use visual annotations~\cite{shtedritski2023does, nasiriany2024pivot, convoi, liu2024moka}, such as lines and numbers, to aid the VLM to effectively comprehend spatial information. Instead of randomly generating the candidates like in PIVOT~\cite{nasiriany2024pivot}, we use a generative model-based trajectory generation approach to produce diverse candidate trajectories that ensure traversability for the VLM to choose from. 


\section{Approach}
\label{sec:approach}

In this section, we formulate the problem of mapless global navigation and describe our approach.

\subsection{Overview}
\label{sec:problem_definition}
Our approach computes a trajectory in a mapless environment for global navigation. {Mapless global navigation requires a robot to reach a distant target beyond its immediate surroundings without relying on a pre-built map. To achieve this, we utilize multi-modal sensor data, combining both geometric and RGB visual information, to iteratively generate local trajectories that guide the robot towards the goal. Our approach follows a two-stage pipeline, as illustrated in Fig.~\ref{fig:architecture}. In the first stage, we generate multiple candidate trajectories, each spanning a fixed length (e.g., 10$\mathrm{m}$) that satisfy the geometric traversability constraints. }
Then, we select the best trajectory based on human-like decision-making. Given a target goal $g\in\co_g$, we use a GPS sensor to provide the relative position between the target and the current location. 
Our goal is to compute a trajectory, $\tau$, that aims to provide the best path to the goal, and that satisfies the traversability constraints of the scenario, $\tau = \textrm{\ours}(\ell, \i, \o, g)$, where $\o=\set{\o_l,\o_v,\i}$ represents the robot's observations. $\o_l\in\co_l$ represents LiDAR observations, $\o_v\in\co_v$ indicates the robot's velocity, and $\i\in\ci$ represents the RGB images from the camera. $\ell \in \cl$ represents the language instructions to the Vision-Language Models (VLMs) for acquiring traversable trajectories. 

We use Conditional Variational Autoencoder {(CVAE)}~\cite{mtg} to process the geometric information, $\o_l\in\co_l$, from the LiDAR sensor and the consecutive velocities, $\o_v\in\co_v$, from the robot's odometer. 
We efficiently generate a set of trajectories lying in geometrically traversable areas, $\ct = \textrm{CVAE}(\o_l,\o_v)$. These generated trajectories cannot handle geometrically similar but color-semantically different situations, such as crosswalks as shown in Fig.~\ref{fig:cover} (C). Therefore, we use VLMs to provide scene understanding from the RGB images.

However, the generated real-world waypoints from CVAE and the image observations are in two different modalities. To fuse these, we overlay the trajectories onto the images. VLMs are then used to assess whether the trajectories align with the contextual constraints of the environment. We assume that VLMs can infer common-sense reasoning from the images. 
We place these numbers at the end of each trajectory, starting from $0$. The numbers indicate the order of distances to the goal, with the lowest number corresponding to the trajectory with the shortest distance. Thus, we map the real-world trajectories to image pixel-level objects by 
{\begin{equation}
    (\n, \ct_c) = M(\textrm{CVAE}(\o_l,\o_v), K),
    \label{eq:selection}
\end{equation}}
where $K$ denotes the conversion matrix from the real-world LiDAR frame to the image plane, $\ct_c$ denotes the converted trajectories, and $\n\in\cn$ are the numbers corresponding to each trajectory. 

Given the language instruction $\ell$, the image $\i$ with the converted trajectories $\ct_c$, and numbers $\n\in\cn$, our VLM selects one traversable trajectory based on the color-semantic understanding of the scenarios:
{\begin{equation}
    \tau = \textrm{VLM}(\ell, \i, \ct_c, \n).
    \label{eq:vlm}
\end{equation}
}
We choose the trajectory with the highest probability as the human-like trajectories, $\max P(\tau | \ell, \i, \ct_c, \n)$. Therefore, the problem is defined as:
\begin{equation}
    {\max\; P(\tau | \ell, \i, \ct_c, \n).}
    \label{eq:max_prob}
\end{equation}



\subsection{Geometry-based Trajectory Generation}

The trajectory set, $\ct$, is generated by a CVAE to generate trajectories with associated confidences. For each observation $\set{\o_l, \o_v}$, we calculate the condition value $\c=f_e(\o_l, \o_v)$ for the CVAE decoder, where $f_e(\cdot)$ denotes the perception encoder. The embedding vector is then calculated from $\c$ as $\z = f_z(\c)$, with $f_z(\cdot)$ representing a neural network. 

To generate a sufficient number of candidates for the robot's navigation, we need to create multiple diverse trajectories that cover all traversable areas in front of the robot. Since the decoder is designed to generate a single trajectory from one embedding vector, producing a variety of diverse trajectories requires the use of representative and varied embedding vectors. We project the embedding vector $\z$ onto orthogonal axes by linear transformations, each projected vector corresponding to one traversable area. Then we generate trajectories based on the condition $\c$:
\begin{equation}
    \z_k = A_k(\c)\z + b_k(\c)= h_{\psi_k}(\z),
    \label{eq:diversity}
\end{equation}
where $h_{\psi_k}$ denotes the linear transformation of $\z$. 
Using each embedding vector $\z_k$, the decoder generates a trajectory $\tau_k$, as $p \left ( \tau_k | \z_k, \c, \Bar{\cz}_k \right )$. $\tau_k\in \ct$ represents generated trajectories. $\z_k$ and $\Bar{\cz}_k$ are the embedding vectors of the current trajectory and the set of other trajectory embeddings, respectively. 
The training of the trajectory generator is the same as MTG~\cite{mtg}, where we use traversability loss, CVAE lower bound, and diversity loss to train the model. 


\subsection{VLM-based Trajectory Selection}


\setlength{\textfloatsep}{0.5cm}
\setlength{\floatsep}{0.5cm}

\begin{algorithm}[tb]
\caption{Multi-modal Trajectory Generation and Selection Algorithm
}\label{alg:adaptive_schedule}
\SetKwInOut{Given}{Given}
\SetKwInOut{Initialize}{Initialize}
\Given{LiDAR point cloud $\o_l$, robot's velocities $\o_v$, transformation matrix $K$, threshold $d_t$, instruction $\ell$, RGB image $\i$}
\Initialize{trajectory set $\ct = \{ \}$, time stamp $t=2$}
\While{the robot is running}{
    $\ct_n = \textrm{CVAE}(\o_l, \o_v)$\;
    $\ct = \ct_n \bigcup \ct$\;
    $(\n, \ct_c) = M(\ct, K)$\; 
    $\tau = \text{VLM}(\ell, \i, \ct_c, \n)$\;
    \uIf{$t_\ct>t$}{ 
        $\ct$.DEQUEUE$()$\;
    }
}
\end{algorithm}


Algorithm~\ref{alg:adaptive_schedule} highlights our procedure of using VLMs to select a suitable trajectory from candidate trajectories. $t_\ct$ denotes the time steps $\ct$ contains. $\ct_n$ denotes a new set of trajectories generated by CVAE. While the generated trajectories $\ct_n$ effectively cover the traversable areas in front of the robot~\cite{mtg}, the deep-learning-based generative model cannot guarantee the consistent generation of traversable trajectories. To address this, we sample consecutive $t=2$ time steps, introducing redundancy to increase the likelihood that at least one of the generated trajectories will be traversable. 
Given the collected trajectories in $\ct$, we convert them to the image plane with numbers, where we sort the trajectories in terms of heuristic, which is the distance between the last waypoint of the trajectory and the goal, {as shown in Eq.~\ref{eq:selection}}.

Considering that trajectories generated at consecutive time steps often overlap significantly, we refine the set of candidates $\ct$. This is done by selecting only representative trajectories to form a subset $\ct'\subseteq \ct$ based on their Hausdorff distances: 
\begin{equation}
    \forall \tau_n, \tau_m \in \ct', \; d_h(\tau_n, \tau_m) > d_t, \;\text{where}\; n\neq m,
\end{equation}
where $d_h(\cdot,\cdot)$ represents the Hausdorff distance. This process removes trajectories that are too similar, improving the clarity of visual annotations on the image while ensuring diversity. 

We then project the trajectories $\ct'$ from the robot's frame to the image plane by transformation matrices K, $\ct_c = P_c(\ct', K)$. 
Following the trajectory generation sequence, we annotate the trajectories with numbers, $\n$.

Finally, we use the VLM to select the best trajectory in terms of satisfying traversability and social compliance. The annotated trajectories $(\n, \ct_c)$ and the current observation image $\i$ are input into the VLM with the prompt instruction $\ell$. The VLM selects the best trajectory, $\tau$, in terms of traversability, social compliance, and traveling distance to the goal, as shown in Eq.~\ref{eq:vlm}. 
{The following is the instruction prompt instruction $\ell$ provided to the model}:
\begin{mybox}
I am a wheeled robot that cannot go over high bumps. This is the image I am seeing right now. \\
Pick one path that I should follow to navigate safely towards the goal, like what humans do. Remember that I must walk on pavements, avoid rough, bumpy terrains, and follow the rules. I cannot go over/under the curbs.\\
The lowest number indicates the shortest path to the goal. Pick only one.\\
Provide the answer in this form: \{`trajectory': []\}
\end{mybox} 

Given the selected trajectory $\tau$, our motion planner generates the corresponding robot action $\a$ to follow it. The VLM is re-prompted each time it returns a response. Although our VLM-based trajectory selector operates at a relatively low frequency, \ie, every 2 to 4 seconds, {the trajectory generator efficiently produces 10$\mathrm{m}$ trajectories}, ensuring the latency remains manageable.

\floatsep 1\baselineskip plus  0.2\baselineskip minus  0.2\baselineskip
\textfloatsep 1.7\baselineskip plus  0.2\baselineskip minus  0.4\baselineskip





\section{Exprimental Results}
\label{sec:experiments}

In this section, we present the implementation details and the experimental results of our approach.  

\begin{figure*}[ht!]
\centering
\begin{tabular}{cccccc}
 & \small{(a) Flower bed} & \small{(b) Curb} & \small{(c) Crosswalk} & \small{(d) Corner} \\
\spheading{Others} & \includegraphics[width=0.24\linewidth]{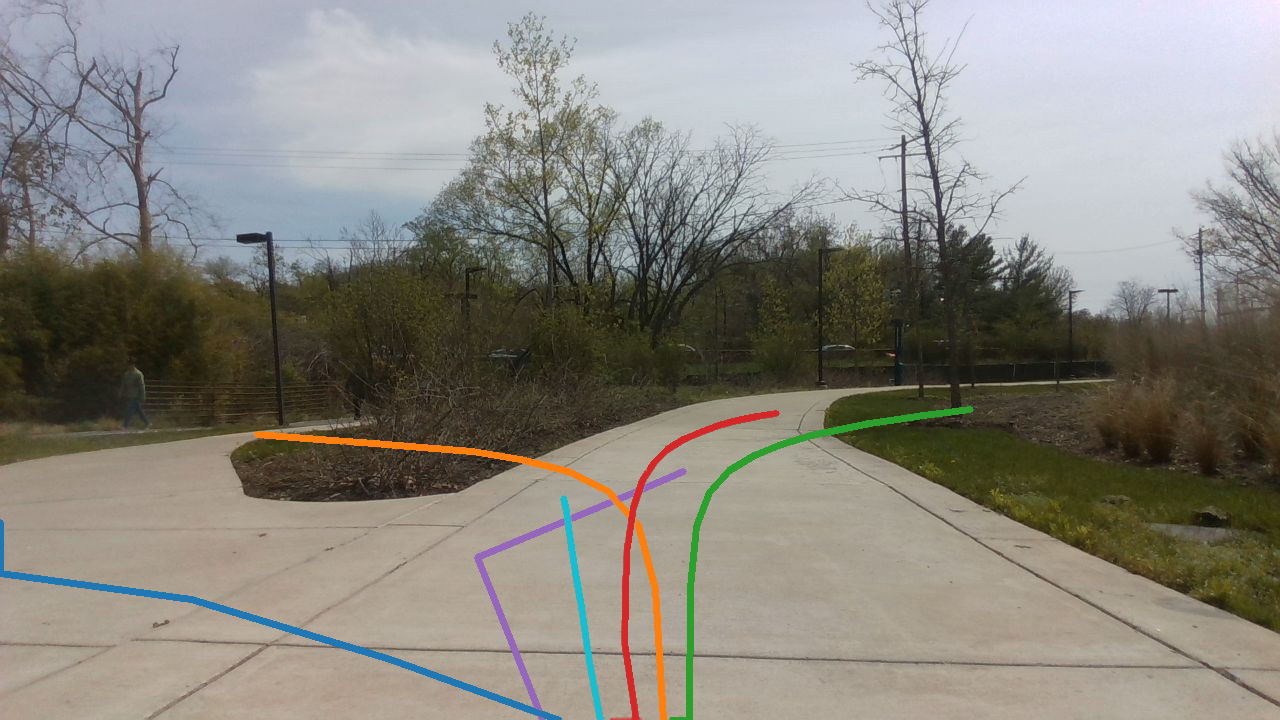} &
\includegraphics[width=0.24\linewidth]{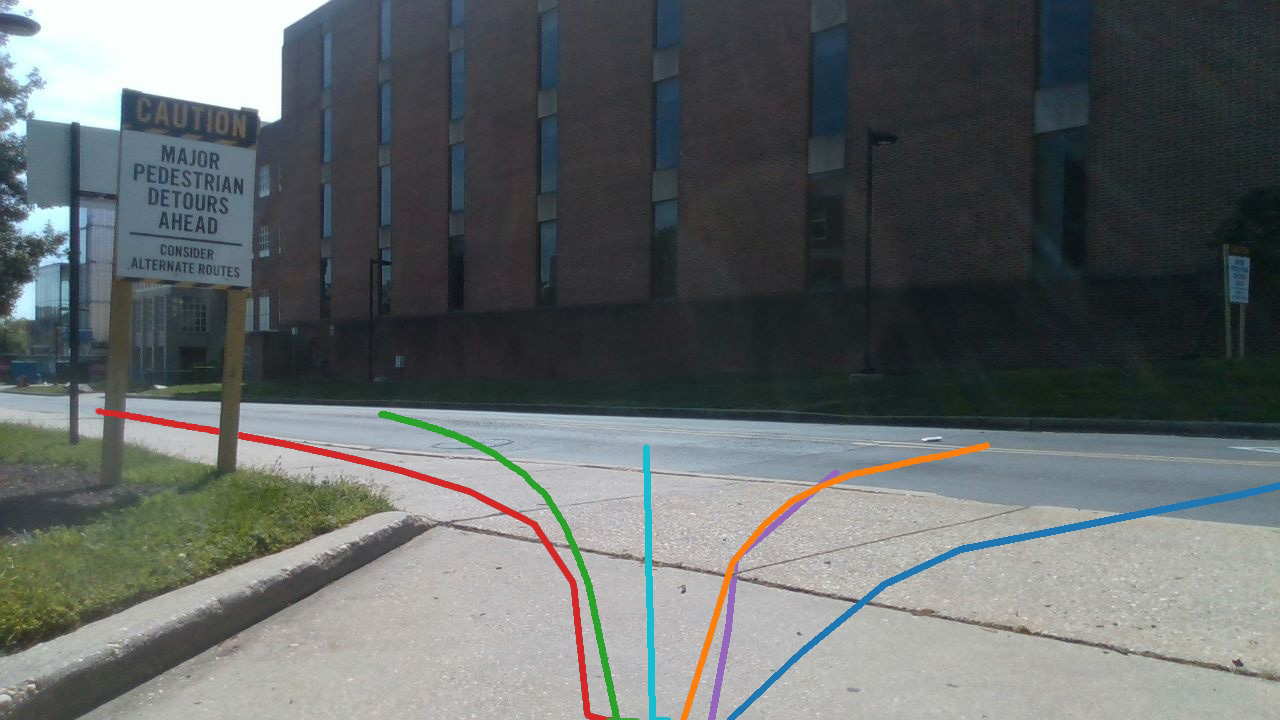} &
\includegraphics[width=0.24\linewidth]{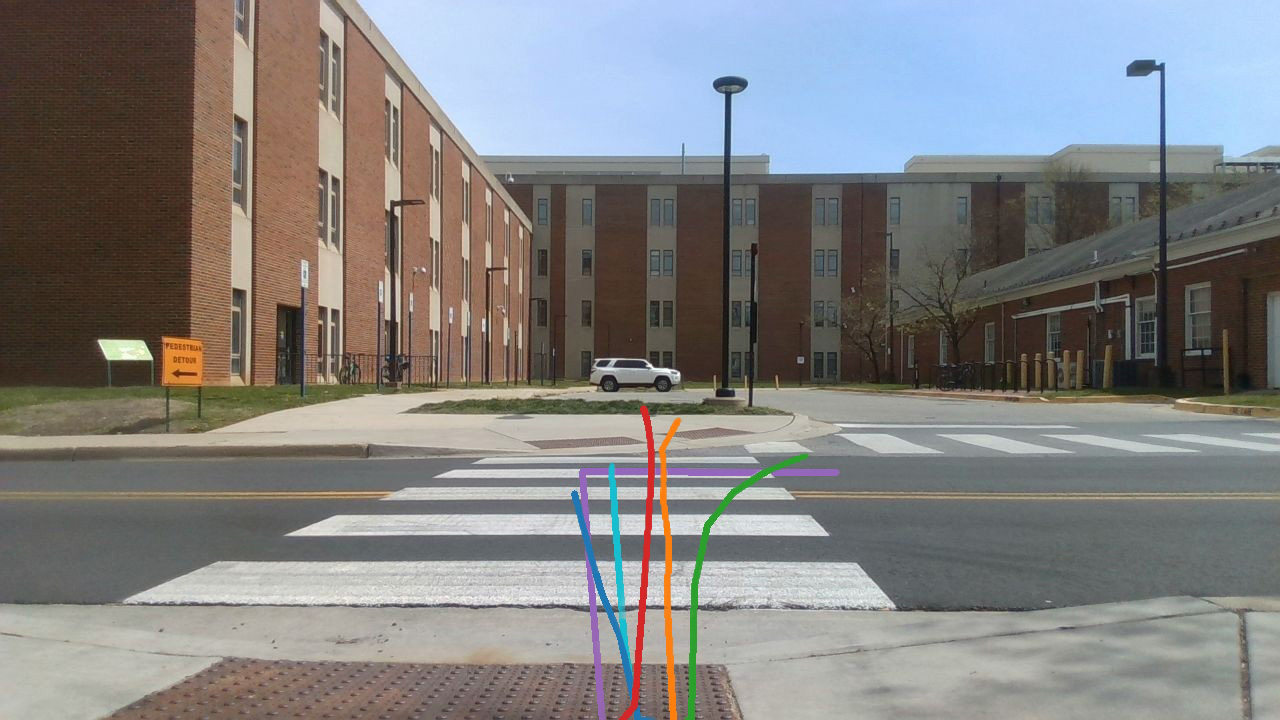} &
\includegraphics[width=0.24\linewidth]{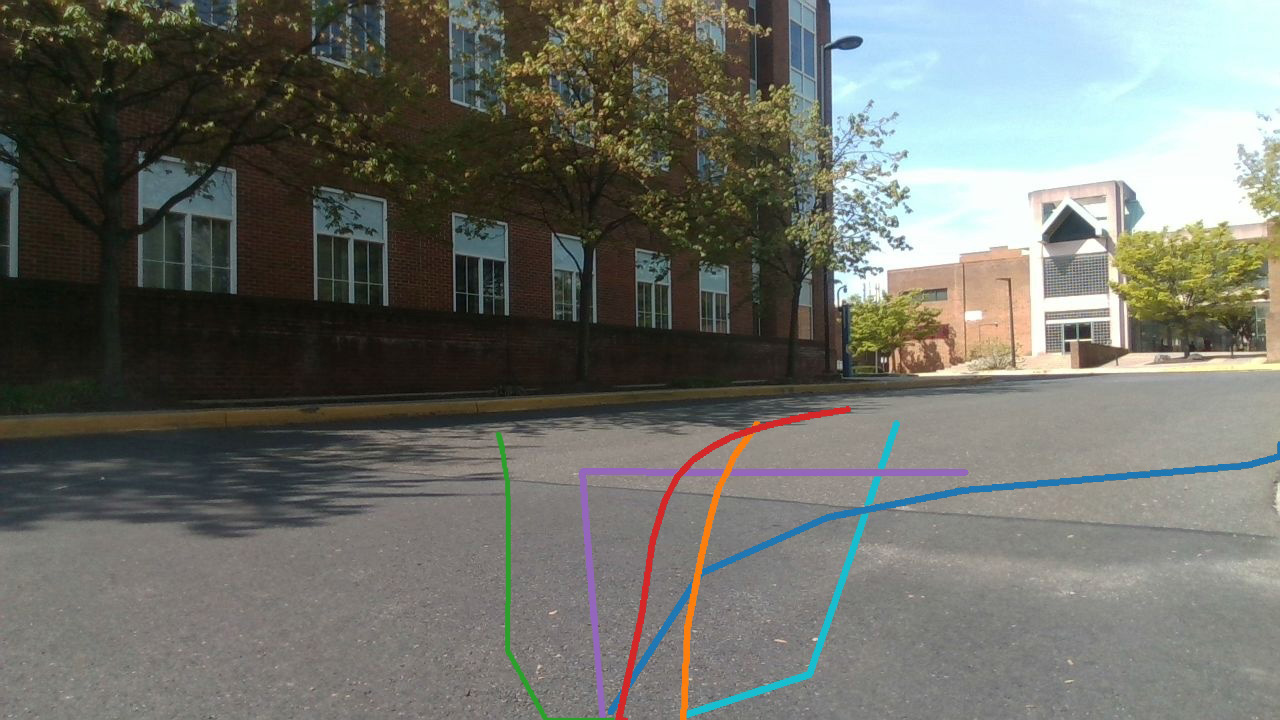} \\

\spheading{\ours} & \includegraphics[width=0.24\linewidth]{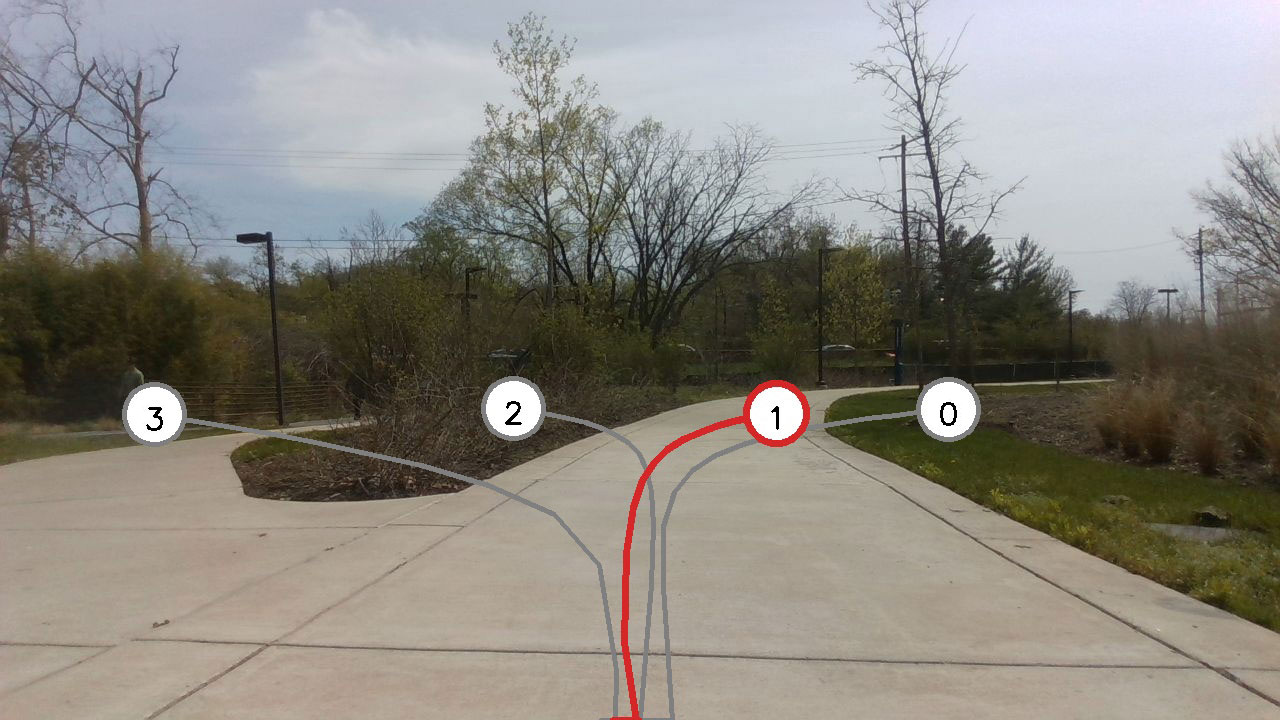} &
\includegraphics[width=0.24\linewidth]{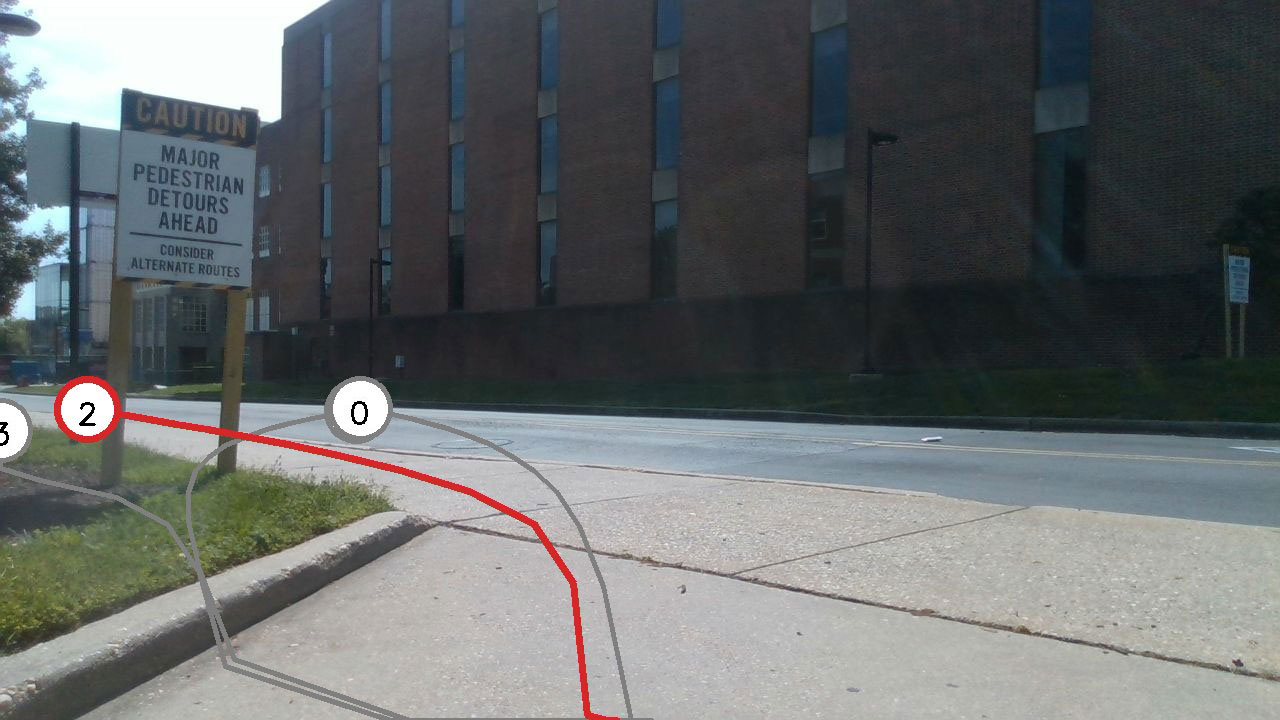} &
\includegraphics[width=0.24\linewidth]{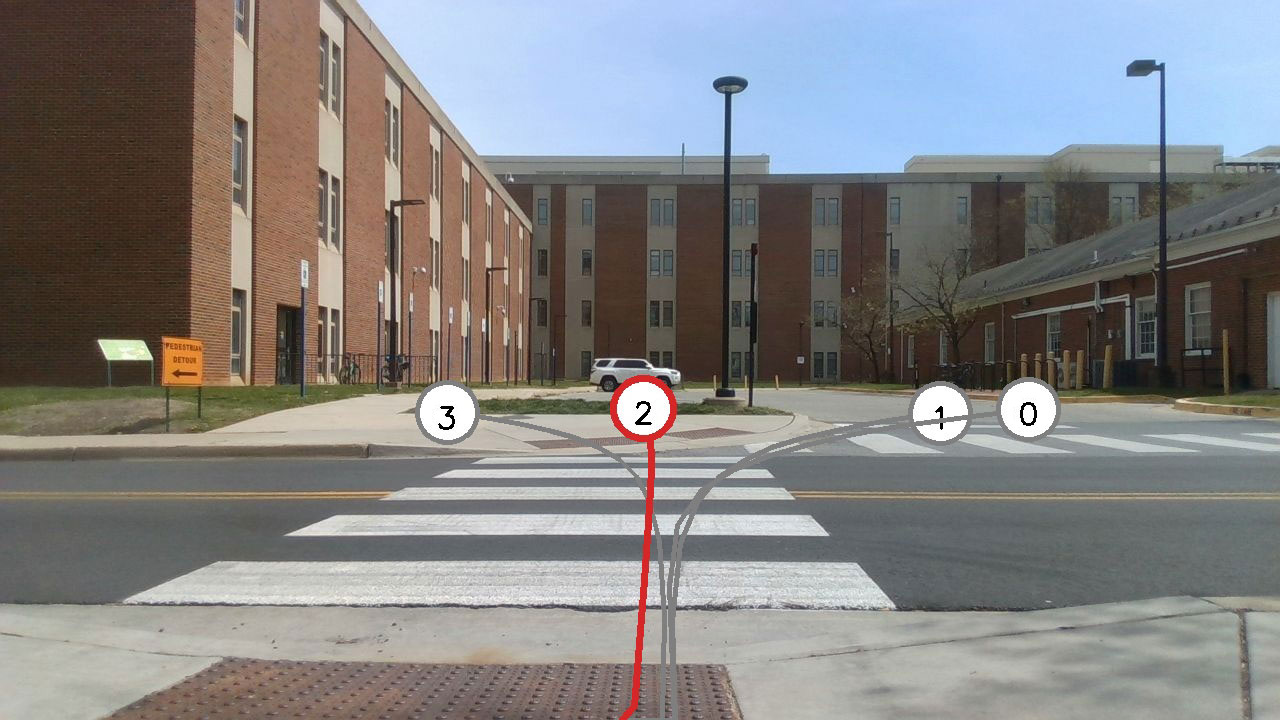} &
\includegraphics[width=0.24\linewidth]{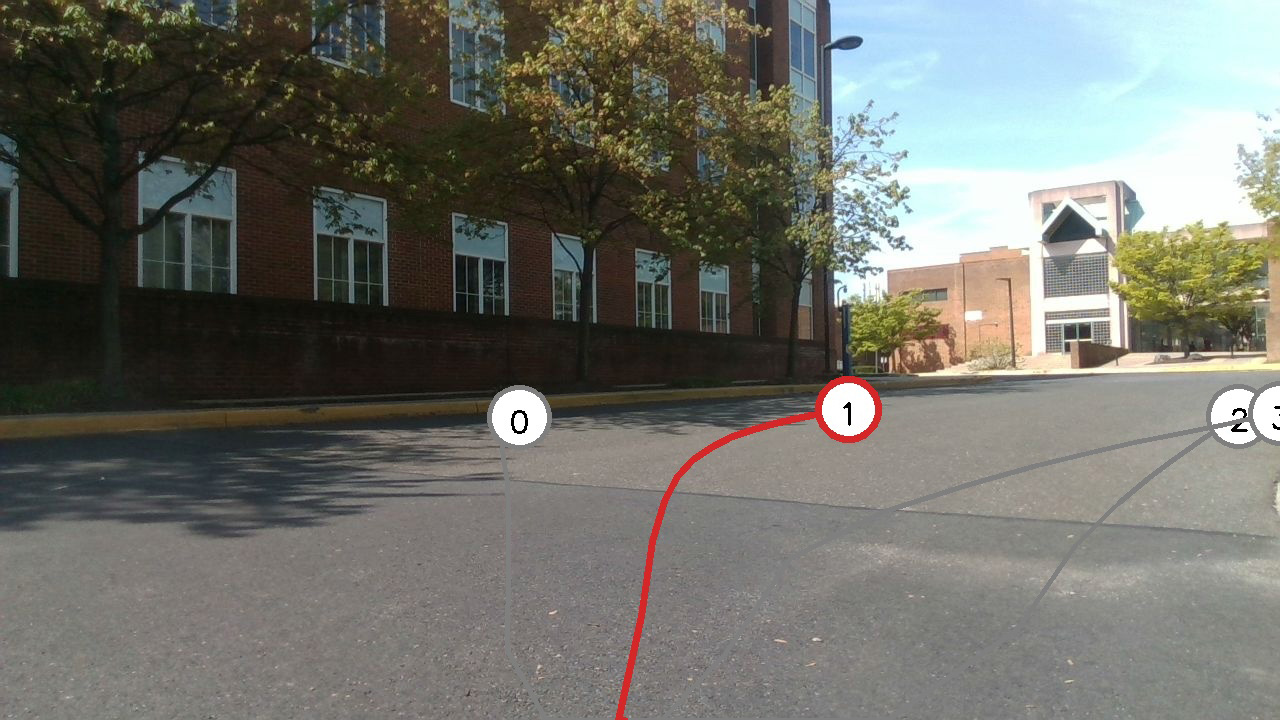} \\

\multicolumn{5}{c}{\small{
\textcolor{mtg}{\thickline} MTG 
 \textcolor{vint}{\thickline} ViNT \textcolor{nomad}{\thickline} NoMaD  \textcolor{pivot}{\thickline} {PIVOT}  \textcolor{convoi}{\thickline} CoNVOI  \textcolor{cand}{\thickline} Candidates 
 \textcolor{ours}{\thickline} \ours}
 }

\end{tabular} 
\caption{\textbf{Qualitative Results:} The top row shows the generated trajectories using all the methods, MTG~\cite{mtg} in green, ViNT~\cite{vint} in blue, NoMaD~\cite{nomad} in orange, {PIVOT~\cite{nasiriany2024pivot} in cyan,} CoNVOI~\cite{convoi} in purple, and \ours~in red. The bottom row shows the candidate trajectories in gray marked with numbers and the selected trajectory in red using \ours. {\ours~can generate and select a trajectory that is both geometrically and semantically feasible.}} 
\label{fig:qualitative}
\vspace{-1em}
\end{figure*}  

\subsection{Implementation Details}

Our approach is tested on a Clearpath Husky equipped with a Velodyne VLP16 LiDAR, a Realsense D435i camera, and a laptop with an Intel i7 CPU and an Nvidia GeForce RTX 2080 GPU. We use CVAE with an attention mechanism to generate multiple trajectories~\cite{mtg} (approximately 10$\mathrm{m}$ each) and use GPT-4V~\cite{achiam2023gpt} 
to select the best traversable trajectory. 

The training dataset~\cite{GND} for our CVAE-based trajectory generation model contains three parts: 1) LiDAR point cloud and robot velocities, 2) binary traversability maps, shown in the right column of Figs. \ref{fig:abla_tg} and \ref{fig:abla_ts}, 3) randomly generated diverse targets with the shortest ground truth trajectories to the targets. 
The binary traversability map is constructed from LiDAR points and is used only for training and evaluation. 
The map is not used during inference. 

To validate \ours, we present qualitative and quantitative results compared with MTG~\cite{mtg}, ViNT~\cite{vint}, NoMaD~\cite{nomad}, {PIVOT~\cite{nasiriany2024pivot},} and CoNVOI~\cite{convoi}. 
We evaluate the performance in four challenging benchmark scenarios: 




  
\begin{itemize}
   \item \textbf{Flower bed:} A robot navigating a paved area next to a flower bed. The robot must stay on the paved path and avoid entering the flower bed. 
  \item \textbf{Curb:} A robot navigating on a sidewalk, which is distinctly separated from the roadway by a curb. The robot must stay on a sidewalk or select a traversable trajectory to go around the curb. 
  \item \textbf{Crosswalk:} A robot crossing the street. The robot must stay on the crosswalk when crossing the street. 
  \item \textbf{Behind the corner:} When the target is behind an obstruction, and there is a large open space ahead, the straight path may lead to an obstacle. The robot must choose a trajectory to navigate around the corner. 
\end{itemize}
{These scenarios pose challenges for navigation without semantic understanding, yet they are common in human-centered environments.}

\subsection{Qualitative Results}

Fig.~\ref{fig:qualitative} shows the resulting robot trajectories corresponding to six different approaches in four different scenarios. The upper row shows the trajectories generated by all the comparison methods including ours and the lower row shows the results of \ours~with the candidate trajectories (gray) and the selected one (red). 

{As MTG relies solely on LiDAR's geometric data, it is unable to deal with traversability differences in flower beds, curbs, and crosswalks, where structure alone provides little distinction. Also, in the corner case where the goal is located around a bend or behind a structure, MTG tends to fail by attempting to cut through rather than effectively navigating around the structure. The performances of ViNT and NoMaD heavily depend on the quality of pre-built topological maps. While they perform well when following straight paths with distinct visual features, such as a crosswalk, they often struggle in environments with turns or significant scene variations. 
While PIVOT selects the most semantically feasible trajectory from the given candidates, it does not explicitly detect geometric information and its random trajectory generation disregards both geometric and semantic information, potentially resulting in no viable options for the VLM to choose from.} Compared to other methods, CoNVOI generally produces trajectories that are both geometrically and semantically feasible. However, its zigzag motion results in non-smooth robot movements. As shown in the bottom row of each scenario in Fig.~\ref{fig:qualitative}, our approach produces diverse trajectories and selects the best one that is traversable and contextually appropriate.


\setlength{\tabcolsep}{6pt}

\begin{table*}[tb] 
\centering
\caption{{\textbf{Quantitative Results:} Comparisons with state-of-art methods}} 
\vspace{-0.5em}
\begin{tabular}{ccccccc} 
 \midrule
 \hspace{0.5em}\multirow{2}{*}{\textbf{Metric}}\hspace{0.5em} & \hspace{0.5em}\multirow{2}{*}{\textbf{Method}}\hspace{0.5em} & \hspace{0.5em}\multirow{2}{*}{\textbf{Input}}\hspace{0.5em} & 
 \multicolumn{4}{c}{\textbf{Scenario}} \\
 \cmidrule(lr){4-7}
  & & & Flower bed & Curb & Crosswalk & Corner \\
 \midrule
 \multirow{6}{*}{\shortstack[c]{Travers-\\ability\\(\%) $\uparrow$}} 
 & MTG~\cite{mtg} & L & 58.19 {$\pm$ 16.65} & 67.12 {$\pm$ 15.65} & 61.82 {$\pm$ 10.95} & 44.71 {$\pm$ 18.35} \\
 & ViNT~\cite{vint}& I & 63.62 {$\pm$ 18.49}& 78.37 {$\pm$ 17.94} & 84.78 {$\pm$ 3.16}  & 44.95 {$\pm$ 17.16} \\
 & NoMaD~\cite{nomad}& I &75.64 {$\pm$ 15.04}& 83.13 {$\pm$ 10.04} & 79.24 {$\pm$ 13.36} & 77.38 {$\pm$ 18.59}\\
 & {PIVOT~\cite{nasiriany2024pivot}} & {I} & {64.75 $\pm$ 19.63} & {79.58 $\pm$ 12.86} & {76.78 $\pm$ 10.41} & {68.66 $\pm$ 15.76} \\
 & CoNVOI~\cite{convoi} & I+L & 81.10 {$\pm$ 9.98} & 75.68 {$\pm$ 12.86} & 86.24 {$\pm$ 12.63} & \textbf{88.46} {$\pm$ 11.45}\\
 & \ours~(Ours) & I+L & \textbf{87.22} {$\pm$ 10.27}& \textbf{89.93} {$\pm$ 7.11} & \textbf{87.44} {$\pm$ 9.78} & 78.00 {$\pm$ 7.79}\\
 \midrule
 \multirow{6}{*}{\shortstack[c]{Fr{\'e}chet\\Distance\\ ($\mathrm{m}$) $\downarrow$}} 
 & MTG~\cite{mtg}  & L & 6.61 {$\pm$ 1.91} & 8.40 {$\pm$ 6.30} & 10.42 {$\pm$ 2.53} & 9.93 {$\pm$ 3.04}\\
 & ViNT~\cite{vint} & I & 10.43 {$\pm$ 2.92} & 10.78 {$\pm$ 3.08} & 8.94 {$\pm$ 2.29} & 12.71 {$\pm$ 2.43}\\
 & NoMaD~\cite{nomad} & I & 7.65 {$\pm$ 3.32} & 8.71 {$\pm$ 3.53} & 11.87 {$\pm$ 2.99} & 9.62 {$\pm$ 2.60}\\
 & {PIVOT~\cite{nasiriany2024pivot}} & {I} & {8.41 $\pm$ 1.85} &  {7.86 $\pm$ 1.55} & {10.53 $\pm$ 3.00} & {9.48 $\pm$ 3.15}\\
 & CoNVOI~\cite{convoi} & I+L & 11.64 {$\pm$ 0.47} & 12.24 {$\pm$ 1.12} & 11.33 {$\pm$ 1.26} & 12.36 {$\pm$ 2.15}\\
 & \ours~(Ours) & I+L & \textbf{5.27} {$\pm$ 1.65} & \textbf{7.93} {$\pm$ 1.28} & \textbf{6.38} {$\pm$ 2.95} & \textbf{8.49} {$\pm$ 2.29}\\
 \midrule
\end{tabular}
\label{table:quantitative}
\vspace{-1.0em}
\end{table*} 

\setlength{\tabcolsep}{1pt}

\subsection{Quantitative Results}
To further validate \ours, we evaluate the methods using two different metrics:
\begin{itemize}
    \item \textbf{Traversability: } The ratio of the generated trajectory lying on a traversable area. The binary traversability map, initially generated using LiDAR, is used for evaluation. This metric is calculated as:
      \begin{align}
        tr(\ca, \hat{\btau}) = 
        \sum_{m=1}^M c(\ca, \w_m), \;\; \w_m \in \hat{\btau},
        \label{eq:traversability}
        \end{align}
        where $c(\cdot, \cdot)$ tells if the waypoint $\w_m$ is in the traversable area $\ca$.
  
    \item \textbf{Fr{\'e}chet Distance w.r.t. Human Tele-operation: } Fr{\'e}chet Distance \cite{alt1995computing} is one of the measures of similarity between two curves. We measure the similarity between the trajectories generated by the methods and the trajectory tele-operated by human.
    A lower distance indicates a higher degree of similarity. 
\end{itemize}

Table~\ref{table:quantitative} reports the results averaged over 20 different frames, with five repetitions for each frame, scenario, and method. Fig.~\ref{fig:qualitative} shows one of the examples. In the Input column, L indicates LiDAR point cloud and I indicates RGB images. While MTG, ViNT, NoMad, and PIVOT rely on a single sensory input, CoNVOI and \ours~utilize both LiDAR point clouds and RGB images. The results demonstrate that \ours~outperforms other state-of-the-art approaches in most of the cases. Specifically, we achieve at least $3.35\%$ and at most $47.74\%$ improvement in terms of average traversability, and at least $19.62\%$ and at most $40.99\%$ improvement in terms of average Fr{\'e}chet distance. 
Overall, the average improvement rates are approximately $20.81\%$ for traversability and $28.51\%$ in Fr{\'e}chet distance. 

In Table~\ref{table:quantitative}, we observe that MTG produces very low results in terms of traversability. This is not only because our benchmark scenarios were selected based on scenarios that are difficult to detect with LiDAR, but also because MTG often fails to consider traversability while focusing on optimality to the goal. In terms of Fr{\'e}chet distance, MTG and \ours~produce good results because they output smooth trajectories similar to a human-operated trajectory we compare against. In contrast, CoNVOI generates a linear trajectory that differs significantly from typical human-operated trajectories, resulting in a lower similarity. 
{CoNVOI generates short trajectories using only two waypoints, reducing the likelihood of waypoints landing in non-traversable areas and leading to a high traversability result. However, in practice, intermediate points may still fall into non-traversable regions. 
Both ViNT and NoMaD are image-based navigation approaches, but NoMaD outperforms ViNT {in 3 out of 4 cases} in terms of traversability and Fr{\'e}chet distance. While both perform well in straight-line following scenarios (e.g., crosswalks), they tend to go off-course when robots are taking turns or the scenarios are dynamic. Additionally, since some of our flower bed and curb scenarios included smooth turns, their variance is notably high. 
As PIVOT generates random straight-line candidates, its performance is inconsistent, exhibiting high variation in results. }
The result demonstrates that \ours~generates human-like trajectories in human-centered environments while ensuring good traversability.

\subsection{Ablation Studies} 

To evaluate the capability of different components of our innovations, we compare \ours~with two different settings. 
First, we compare by removing our CVAE-based trajectory generator. Instead, we randomly generate trajectories. This approach aligns with the method utilized by PIVOT, but we omit their iterative questioning mechanism as part of our ablation study.
Second, we compare by removing our VLM-based trajectory selector. Instead, we select a trajectory by using a heuristic to select the shortest travel distance to the goal, which aligns with the approach, MTG. 

\begin{figure}[tb]
\centering
\begin{tabular}{ccc} 

    \spheading{(a) CoNVOI} & \includegraphics[height=6.5em]{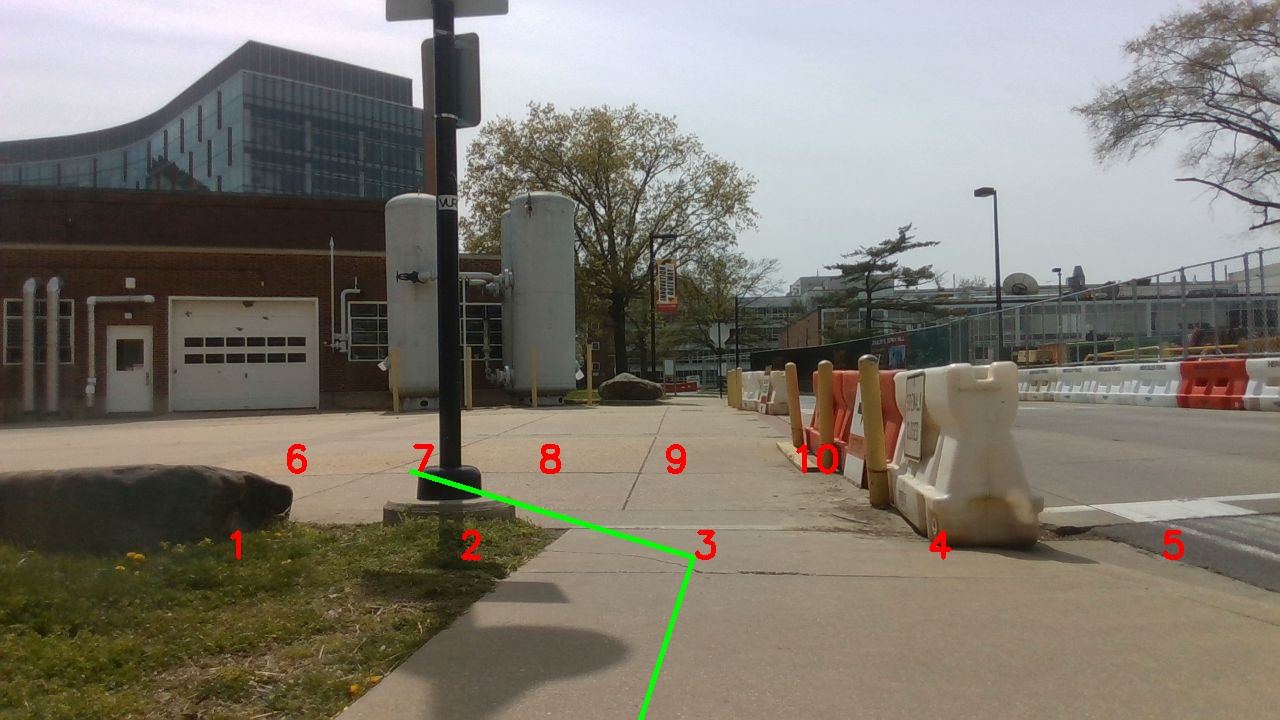} &
    \includegraphics[height=6.5em]{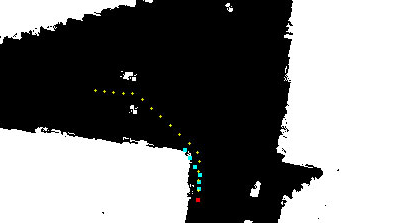}\\
    \spheading{(b) PIVOT} & \includegraphics[height=6.5em]{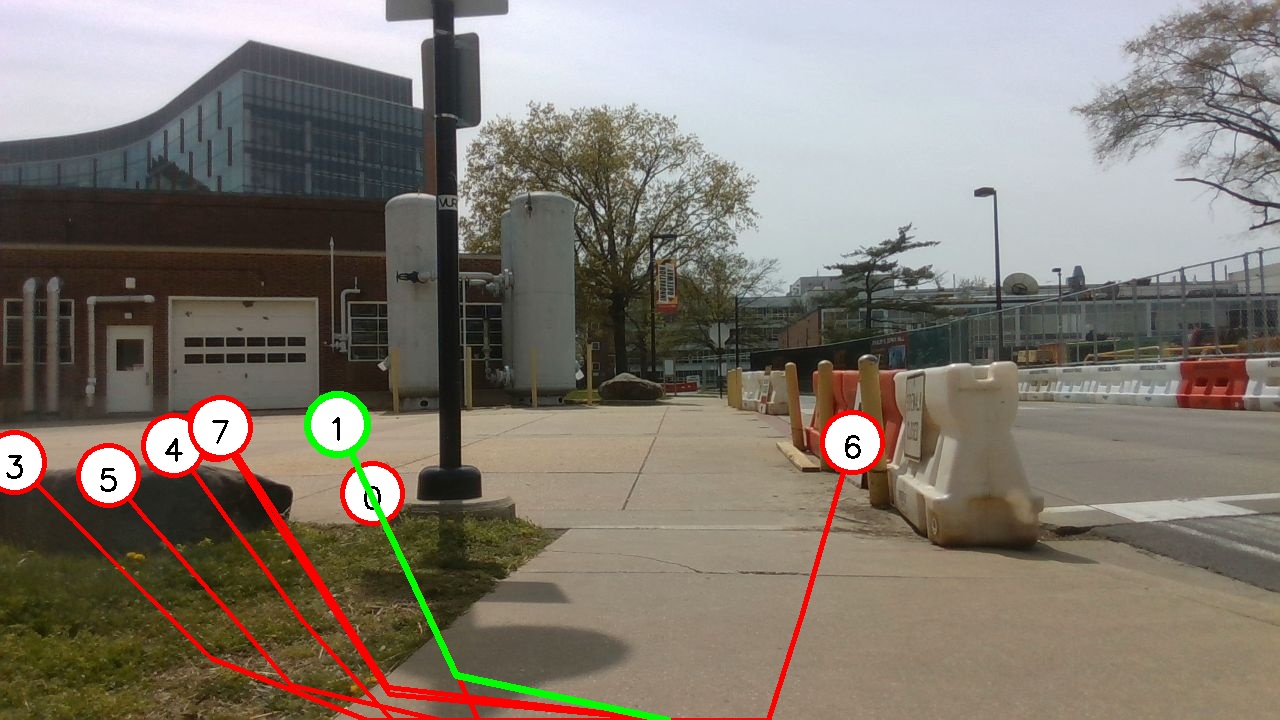} &
    \includegraphics[height=6.5em]{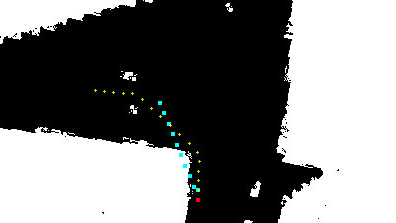}\\
    \spheading{(c) \ours} & \includegraphics[height=6.5em]{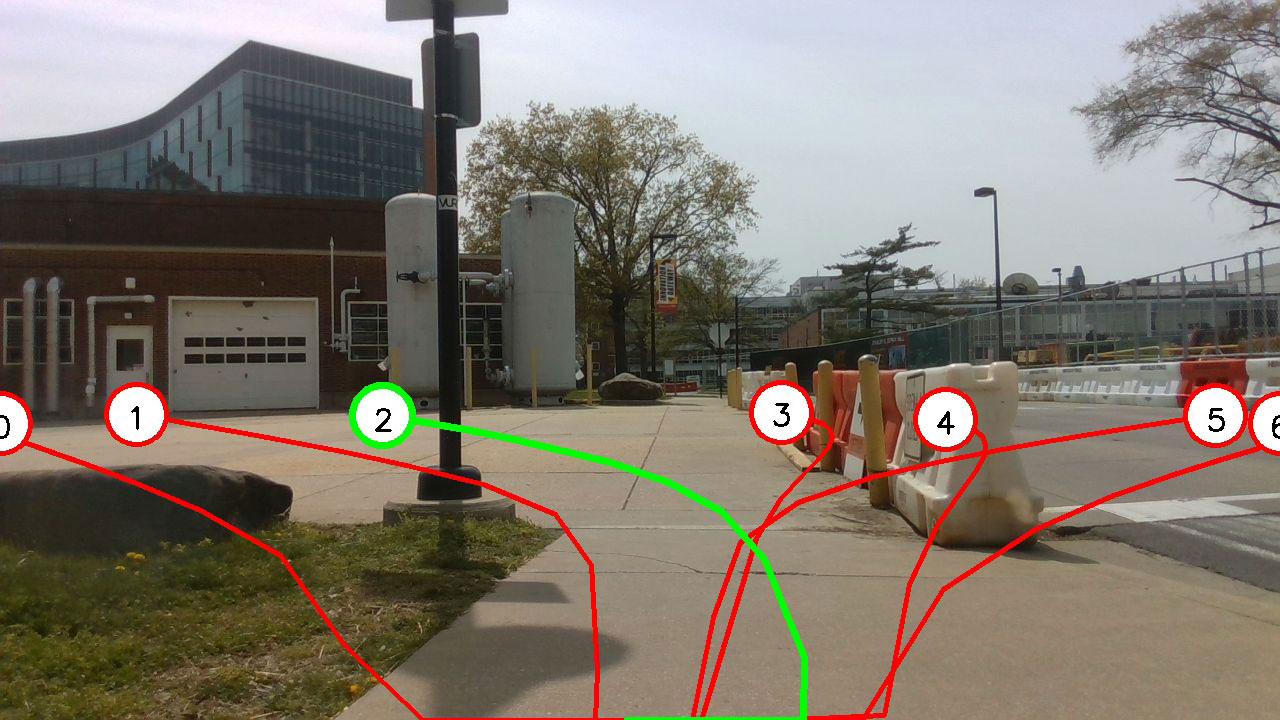} &
    \includegraphics[height=6.5em]{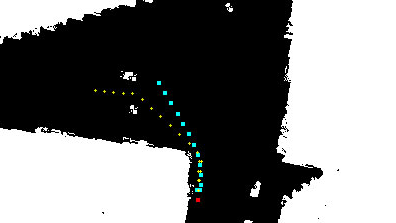}\\
    \multicolumn{3}{c}{\small{
    \textcolor{red}{\thickline} Candidates 
 \textcolor{green}{\thickline} Final Trajectory}}
    
\end{tabular} 
\caption{\textbf{Ablation Study on the Trajectory Generator:} 
The left shows the generated candidate trajectories (red) and the selected trajectory (green) in the robot-view image. The right shows the top-down view image of the traversability map. The cyan color represents the final selected trajectory, and the yellow color represents the human-driven trajectory. Compared with CoNVOI~\cite{convoi} and PIVOT~\cite{nasiriany2024pivot}, \ours~generates the trajectory closest to the human-driven one, which keeps the robot on a safe pavement surface.} 
\label{fig:abla_tg}
\end{figure}

\textbf{Ablation on Trajectory Generator: } Fig.~\ref{fig:abla_tg} illustrates the ablation study to evaluate the effectiveness of our CVAE-based trajectory generator. 
The red lines and numbers are the inputs given to the VLM. The green line indicates the selected trajectory by the VLM.
PIVOT randomly generates the sub-goal targets and linearly connects them. We randomly generate 10 endpoints that are within 5$\mathrm{m}$ to 15$\mathrm{m}$ ahead and then linearly connect the points to generate trajectories. 
It represents the approach of a VLM-based trajectory selector without a CVAE-based trajectory generator. Because the target is randomly generated, it often fails to generate good candidate trajectories. 
We also compare with CoNVOI, which adopts a different approach to generating candidates for the VLM. CoNVOI marks obstacle-free regions with numerical labels, employs the VLM to select suitable labels, and connects them with straight lines to form a trajectory. However, while the marked regions are obstacle-free, this method does not consider the waypoints between the labels. Consequently, the generated trajectories may intersect obstacles, as demonstrated in Fig.~\ref{fig:abla_tg}(b).
\ours~utilizes the strengths of the CVAE-based trajectory generator to produce high-quality candidates for the VLM to evaluate and select from. 
The study highlights the critical role of having high-quality candidate trajectories, emphasizing the significance of an effective trajectory generator.

\begin{figure}[tb]
\centering
\begin{tabular}{ccc} 
    \spheading{(a) MTG} & \includegraphics[height=6.5em]{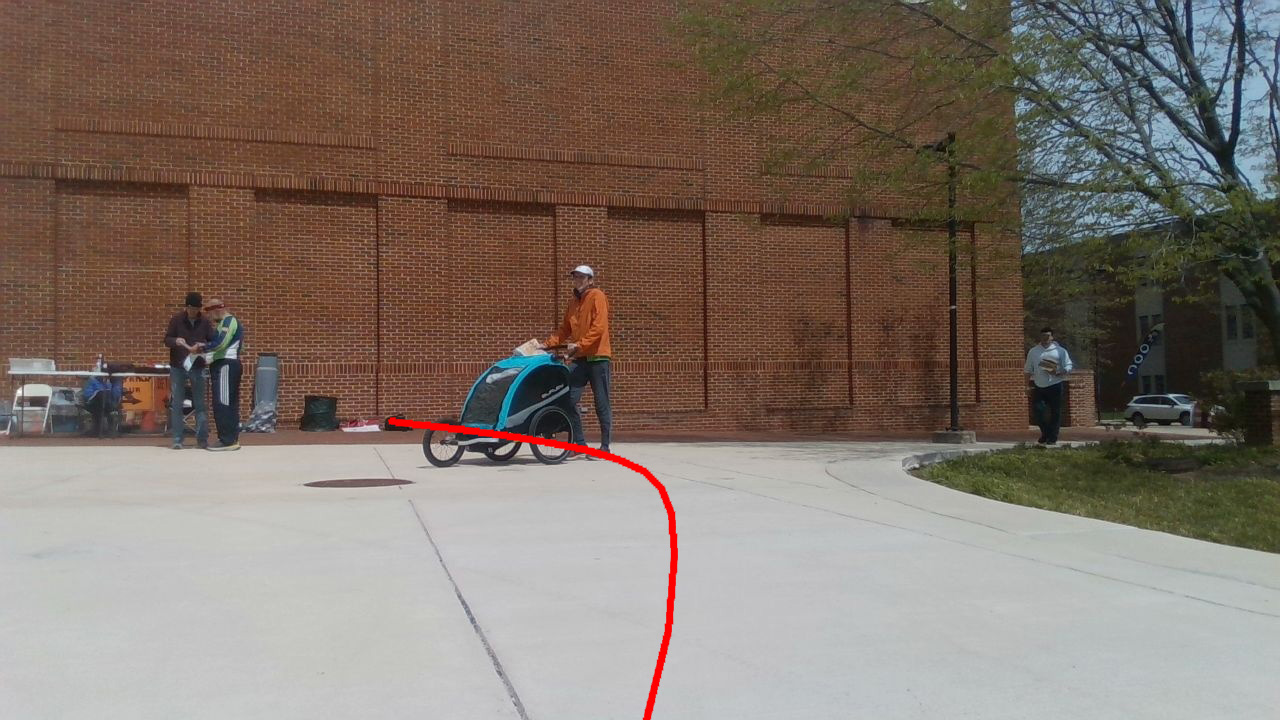} &
    \includegraphics[height=6.5em]{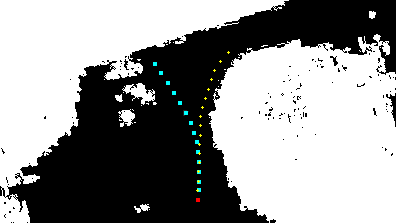}\\
    \spheading{(b) \ours} & \includegraphics[height=6.5em]{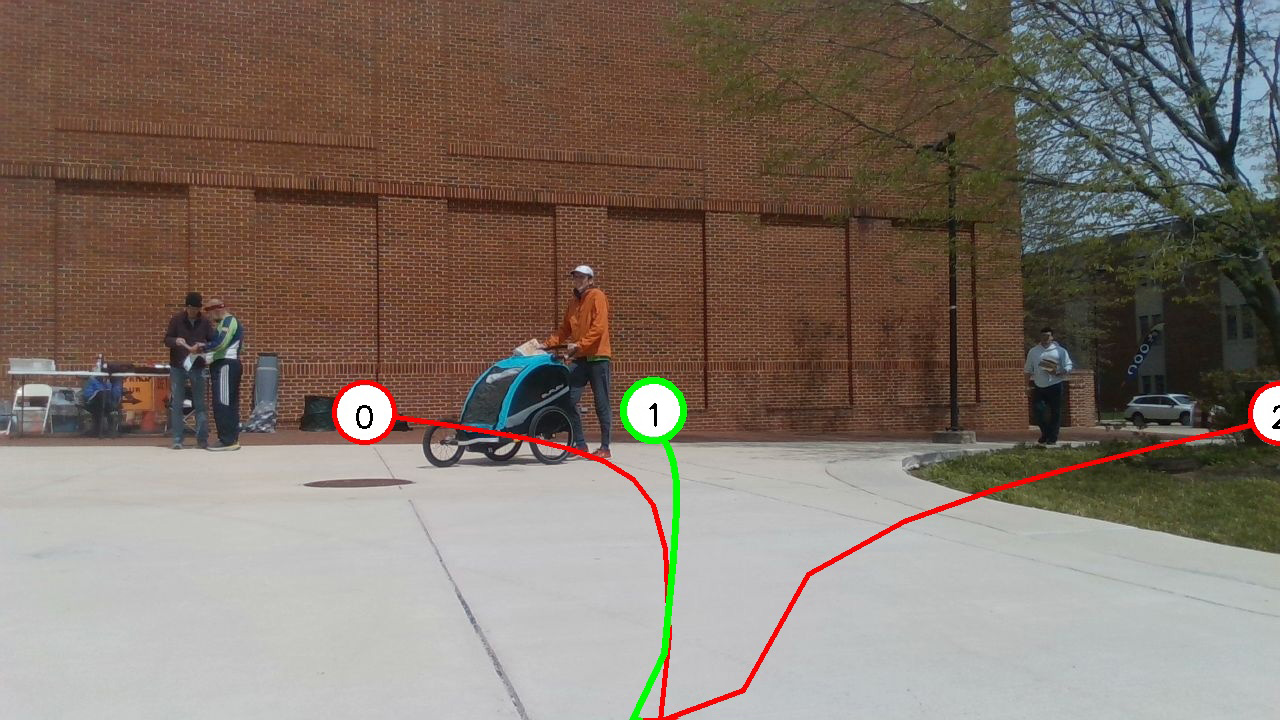} &
    \includegraphics[height=6.5em]{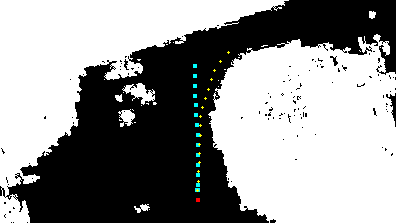}\\

    \multicolumn{3}{c}{\small{
    \textcolor{red}{\thickline} Candidates 
 \textcolor{green}{\thickline} Final Trajectory}} 
 
\end{tabular} 
\caption{\textbf{Ablation Study on the Trajectory Selector:} Compared with MTG~\cite{mtg}, which selects trajectories based on the shortest distance heuristic, \ours~selects the trajectory closer to human-like decision-making, going around the large obstruction.} \vspace{-1.0em}
\label{fig:abla_ts}
\end{figure}

\textbf{Ablation on Trajectory Selector: } Fig.~\ref{fig:abla_ts} illustrates the ablation study to evaluate the effectiveness of our VLM-based trajectory selector. 
MTG uses a CVAE-based approach to generate multiple trajectories and select the optimized trajectory based on a heuristic, the distance to the goal. It represents a CVAE-based trajectory generator without a VLM-based trajectory selector. 
When comparing MTG and \ours, MTG generates a traversable trajectory but often overlooks small, dynamic obstacles such as humans. Additionally, when the target goal is located behind a building, MTG attempts to cut through the building, generating the shortest trajectory to the goal, whereas \ours~selects a trajectory that appropriately navigates around it. The study highlights the importance of the trajectory selector. Rather than relying on a heuristic to choose from candidate trajectories, our VLM-based trajectory selector enables human-like decision-making, driven by the robot's visual perception of the environment.



\subsection{{Real Robot Experiment}}
{In order to demonstrate our approach in the real world, we performed experiments in the real world. 
Fig.~\ref{fig:cover} and Table~\ref{tab:real_exp} show the result of our robot experiments, showcasing a navigation task that incorporates all four scenarios.} {Each method was evaluated in a single run, and failures were counted whenever human intervention was required to recover the robot.} The supplementary video further highlights the resulting robot motions and compares them with other methods.

{In our real robot experiments, we use GPS data to determine both the current robot position and the target goal, which is located approximately 100$\mathrm{m}$ away, positioned behind a building obstruction. To navigate toward the goal, our approach continuously generates 10$\mathrm{m}$ trajectories in a recursive manner. These generated trajectories are then executed using the Dynamic Window Approach (DWA)~\cite{fox1997dynamic}, ensuring smooth and adaptive motion planning. We compare our method against three alternative approaches: MTG, NoMaD, and CoNVOI. 
\ours~exhibits the least number of failures while achieving the shortest travel distance and time.
}

\setlength{\textfloatsep}{0.5em}
\setlength{\floatsep}{0.5em}

\setlength{\tabcolsep}{6pt}
\begin{table}[tb]%
\vspace{-0.5em}
\centering
  \caption{{\textbf{Real World Experiment Results}}}\label{tab:real_exp}
    \begin{tabular}{cccc}
    \toprule
    \multirow{2}{*}{\textbf{Method}} & \textbf{Number of} & \textbf{Travel} & \textbf{Travel} \\  
    & \textbf{Failures} $\downarrow$ & \textbf{Distance ($\mathrm{m}$)} $\downarrow$ & \textbf{Time (sec.)} $\downarrow$ \\
    \midrule
    MTG~\cite{mtg}     & 8  & 111.65 & 201  \\        
    NoMaD~\cite{nomad}   & 6  & 100.79 & 168  \\  
    CoNVOI~\cite{convoi}  & 13 & 115.77 & 257 \\  
    \ours~(Ours) &  \textbf{4} & \textbf{97.53} & \textbf{151} \\  
    \bottomrule
    \end{tabular} 
\end{table} 
\setlength{\tabcolsep}{1pt}

\floatsep 1\baselineskip plus  0.2\baselineskip minus  0.2\baselineskip
\textfloatsep 1.7\baselineskip plus  0.2\baselineskip minus  0.4\baselineskip

\subsection{Discussions} \label{sec:discussion}
\textbf{Low Frequency of Online Large VLMs: }
A notable limitation of using large VLMs for navigation is their relatively low operational frequency, with outputs typically taking 2 to 4 seconds in our experiments. This latency makes them unsuitable for high-frequency real-time decision-making. However, our approach mitigates this issue effectively by generating relatively long trajectories of approximately 10$\mathrm{m}$, reducing the need for frequent updates. This allows the robot to continue moving smoothly without waiting for constant re-evaluation. 
Additionally, the motion planner ensures the robot continues to follow the selected trajectory while waiting for the VLM's decision. This design allows us to leverage the VLM's contextual reasoning capabilities without compromising navigation reliability. Furthermore, the improvements in VLM processing speed could further enhance system responsiveness. 

\textbf{More Challenging Scenarios: } Although our benchmark scenarios focus on stationary environments, our approach is capable of handling dynamic scenarios involving moving obstacles. As illustrated in Fig.~\ref{fig:abla_ts}, our CVAE-based trajectory generator produces traversable candidate trajectories when evaluated against a traversability map. However, it often overlooks small, dynamic obstacles, such as humans. To complement this, our VLM-based trajectory selector incorporates such factors to identify and select feasible trajectories. Furthermore, while the \ours~module generates and selects a trajectory, the underlying motion planner ensures that the robot adheres to it while dynamically adjusting its motion in response to unexpected obstacles in real time. We employ the DWA as the motion planner for our experiments, but this can be replaced with any other local planning algorithm. In this paper, our primary focus is to demonstrate that VLM can effectively handle human-centered environments that require contextual understanding, such as pedestrian walkways and crossings, ensuring that navigation decisions align with social and environmental cues. We establish our benchmark to reflect these challenges.

\section{Conclusion, Limitations, and Future Work}
\label{sec:conclusion}
We propose \ours, a novel multi-modal Trajectory Generation and Selection approach for mapless outdoor navigation. \ours~integrates a CVAE-based trajectory generation method with a VLM-based trajectory selection process to compute geometrically and semantically feasible, human-like trajectories in human-centered outdoor environments. Our approach achieves a $20.81\%$ improvement in traversability and a $28.51\%$ improvement in similarity to human-operated trajectories on average. 

Our method has a few limitations. Since \ours~relies on VLM, its performance can depend on the robustness of the VLM. However, with the ongoing improvements in VLM technology, it is expected that the robustness of our approach will also improve. Furthermore, our trajectory generation method can be substituted with more advanced approaches in the future, offering the potential for further performance enhancements.


\bibliographystyle{IEEEtran}
\bibliography{IEEEabrv,ref}


\end{document}